\documentclass{article}

\usepackage{hhline}
\usepackage{amsmath}
\usepackage{algorithm}
\usepackage{algorithmic}
\usepackage{graphicx}
\usepackage{import}

\PassOptionsToPackage{round}{natbib}
\usepackage[final]{neurips_2023}

\usepackage[utf8]{inputenc} % allow utf-8 input
\usepackage[T1]{fontenc}    % use 8-bit T1 fonts
\usepackage{hyperref}       % hyperlinks
\usepackage{url}            % simple URL typesetting
\usepackage{booktabs}       % professional-quality tables
\usepackage{amsfonts}       % blackboard math symbols
\usepackage{nicefrac}       % compact symbols for 1/2, etc.
\usepackage{microtype}      % microtypography
\usepackage{xcolor}         % colors

\title{DiffClone: Enhanced Behaviour Cloning in Robotics with Diffusion-Driven Policy Learning}

\author{%
  Sabariswaran Mani\\
  Department of E\&ECE\\
  IIT Kharagpur, India\\
  \texttt{sabaris.offl@kgpian.iitkgp.ac.in} \\
  \And
  Sreyas Venkataraman* \\
  Department of Mathematics\\
  IIT Kharagpur, India\\
  \texttt{vsreyas20@kgpian.iitkgp.ac.in} \\
  \And
  Abhranil Chandra* \\
  David R. Cheriton School of Computer Science\\
University of Waterloo, Waterloo, Canada\\
  \texttt{abhranil.chandra@uwaterloo.ca} \\
  \And
  Yash Sirvi* \\
  Department of CSE\\
  IIT Kharagpur, India\\
  \texttt{yashsirvi@kgpian.iitkgp.ac.in} \\
  \And
  Adyan Rizvi* \\
  Department of Mathematics\\
  IIT Kharagpur, India\\
  \texttt{adyan2004@kgpian.iitkgp.ac.in} \\
  \And
  Soumojit Bhattacharya* \\
  Department of E\&ECE\\
  IIT Kharagpur, India\\
  \texttt{soumojit048@kgpian.iitkgp.ac.in} \\
  \And
    Aritra Hazra \\
  Department of CSE\\
  IIT Kharagpur, India\\
  \texttt{aritrah@cse.iitkgp.ac.in} \\
}

\begin{document}

\pagestyle{empty}
\maketitle

\begin{abstract}

Robot learning tasks are extremely compute-intensive and hardware-specific. Thus the avenues of tackling these challenges, using a diverse dataset of offline demonstrations that can be used to train robot manipulation agents, is very appealing. The Train-Offline-Test-Online (TOTO) Benchmark provides a well-curated open-source dataset for offline training comprised mostly of expert data and also benchmark scores of the common offline-RL and behaviour cloning agents. In this paper, we introduce \textbf{DiffClone}, an offline algorithm of enhanced behaviour cloning agent with diffusion-based policy learning, and measured the efficacy of our method on real online physical robots at test time. This is also our official submission to the Train-Offline-Test-Online (TOTO) Benchmark Challenge organized at NeurIPS 2023. We experimented with both pre-trained visual representation and agent policies. In our experiments, we find that MOCO finetuned ResNet50 performs the best in comparison to other finetuned representations. Goal state conditioning and mapping to transitions resulted in a minute increase in the success rate and mean-reward.  As for the agent policy, we developed DiffClone, a behaviour cloning agent improved using conditional diffusion.

\end{abstract}

\section{Introduction}
Gathering data from robots to learn manipulation policies is typically expensive and time-consuming. Nonetheless, by utilizing pre-collected data that is readily accessible, these concerns may be mitigated. Offline data from various robotics hardware, increases the diversity of the dataset and also its size as more data leads to better training of the current models. This improves the efficacy of the learning techniques and improves generalization and robustness upon transferring the offline policies to online real-world situations. Algorithms like Implicit Q-Learning[\cite{IQL}], Decision Transformers[\cite{DT}], TD3-BC[\cite{fujimoto2021minimalist}] have shown success in solving offline RL problems. Offline benchmark datasets enable and make robotics research more accessible, and helps gauge and compare, results and metrics fairly as was already the case in computer vision and NLP [\cite{imagenet} \cite{glue} \cite{rb2}]. 

The \href{https://toto-benchmark.org/}{Train Offline, Test Online Benchmark} provides a robust offline robot learning dataset to promote equitable research and a solid benchmark to compare the advancement of offline algorithms and visual representations that can effectively utilize diverse real-world data fairly. 

The task at hand is to solve planning tasks involving complex manipulation like pouring and scooping on a Franka-Emica Panda arm using previously collected datasets consisting of over 1.26 million images of robot actions in 1895 trajectories of scooping data and 1003 trajectories of pouring data. The dataset consists of RGB and depth images, along with the joint states of the arm, the actions, and a sparse reward for each time step [\cite{TOTO}]. We leverage this offline data to train a visual representation model and an agent policy for each of the tasks without any on-policy data gathering and fine-tuning. The agents are then tested in a real-robot setup where they are evaluated on their ability to generalize and adapt to previously unseen objects, positions, and other such out-of-distribution settings.

Leveraging this dataset with particularly high-quality expert demonstrations, we propose a framework to improve vanilla behavior cloning agents using diffusion policy. We name our agent DiffClone (\ref{alg:DiffClone}): Enhanced Behaviour Cloning in Robotics with Diffusion-Driven Policy Learning.

In DiffClone, we start by selectively sub-sampling trajectories to create a sub-set of "expert" data. This involves choosing trajectories with the highest rewards, ensuring the dataset captures optimal behaviour. Following this, we employ a Momentum Contrast (MoCo) model, fine-tuned on our datasets, as our visual-encoder backbone. This model processes images to extract relevant states. Once these states are obtained, we normalize them across the dataset to enhance the stability of the policy we intend to learn. Finally, we implement a behaviour cloning agent using a CNN-based Diffusion Policy [\cite{unet} \cite{diffpolicy}]. We chose this strategy over other offline RL alternatives, motivated by our success in generating an expert dataset that accurately represents the distribution of the given trajectories. The "expert" dataset's quality and representativeness allowed us to use behaviour cloning techniques that more effectively and accurately replicated the desired optimal behaviour and policy [\cite{kumar2022prefer}]. 

The rest of the paper is organized as follows, Section 2 elaborates on the background of the problem statement by breaking it down into two sub-problems, and discusses the prevalent methods for solving them. Section 3 introduces our approach to the problem, and provides a detailed step-by-step explanation. Section 4 documents our experiments and the results obtained. Finally, we conclude in Section 5.

\section{Background and Preliminaries}
\label{headings}
We breakdown our approach into two parts namely the pre-trained visual encoder backbone, and the RL/BC method which we use as a decision-making agent. In the next subsections, we will look at an overview of the methods benchmarked in TOTO [\cite{TOTO}] and then proceed to describe our method which improves upon the baseline.

\subsection{Visual Encoders}
One of the core problems that arise when learning a policy from images is the difficulty in learning due to the high dimensions of the image. One of the solutions to this is to train an end-to-end model which learns good representations for the task at hand, but this method requires a considerable amount of data and compute, and may not always be feasible. Thus, we often use pre-trained representations trained on a large dataset such as ImageNet, which reduces dimensionality without much loss in important task-related information. We can then fine-tune them on our dataset, or utilise them in a zero-shot manner. The following subsections describe the pre-trained representations we have considered and utilised.
\subsubsection{Bootstrap Your Own Latent (BYOL)}
Bootstrap Your Own Latent (BYOL) \cite{BYoL} is a self-supervised learning approach aimed at learning representation $y_\theta$ which can be used for downstream tasks. The method involves two neural networks: the online and target networks. The online network is defined by a set of weights $\theta$ and is comprised of three stages: an encoder $f_\theta$, a projector $g_\theta$, and a predictor $q_\theta$. The target network has the same architecture as the online network but uses a different set of weights $\xi$. The target network provides the regression targets to train the online network, and its parameters $\xi$ are an exponential moving average of the online parameters $\theta$.

The main mathematical equation of BYOL is the loss function, which minimizes a similarity loss between $q_\theta(z_\theta)$ and $sg(z'_{\xi})$, where $\theta$ are the trained weights, $\xi$ are an exponential moving average of $\theta$, and $sg$ means stop-gradient. In a more simplified form, the BYoL objective can be written as:

\[ L_{BYOL}(\theta) = E[Z_\theta P_\theta - Z'_{\xi}]^2_F \]

where $Z_\theta$ and $Z'_{\xi}$ are the representations of the online and target networks respectively, $P_\theta$ is the predictor of the online network, and $E$ denotes the expectation \cite{richemond2023edge}. 

At the end of training, everything but $f_\theta$ is discarded, and $y_\theta$ is used as the image representation.
This method is benchmarked using a Resnet-50 architecture trained on the ImageNet dataset as the encoder.

\subsubsection{Momentum Contrast (MoCo)}
Momentum Contrast (MoCo) \cite{MoCo} too is a self-supervised learning algorithm that uses a contrastive loss for learning visual representations. MoCo frames the unsupervised learning process as a dictionary look-up task from recent mini-batches. Each image or view is assigned a key, represented by an encoder network. The learning process trains encoders to perform dictionary look-up, where an encoded "query" should be similar to its matching key and dissimilar to others. Within each mini-batch, one image is treated as a positive sample, while the others serve as negative samples.

The dictionary in MoCo is maintained as a queue of data samples. The encoded representations of the current mini-batch are enqueued, and the oldest are dequeued. This approach decouples the dictionary size from the mini-batch size, allowing the dictionary to be large. Moreover, as the dictionary keys come from the preceding several mini-batches, a slowly progressing key encoder is proposed. This is implemented as a momentum-based moving average of the query encoder, ensuring that the encoder network gradually updates itself over time by taking a weighted average of its current state and its previous state.

The core objective of MoCo is to minimize the Euclidean distance between the same data points in query and key encodings while maximizing the distance between all different data points. This approach effectively enhances the learning of distinct and robust visual features from unlabeled data. The contrastive loss used in MoCo is based on the InfoNCE loss \cite{InfoNCE}, which is formulated as:
\begin{equation}
 L_{MoCo} = -\frac{1}{N}\sum_{i=1}^{N}\log\frac{\exp(q_i^T k_{i+} / \tau)}{\exp(q_i^T k_{i+} / \tau) + \sum_{j=1}^{K}\exp(q_i^T k_{i-}^j / \tau)} 
\end{equation}
where $q_i$ is the query representation, $k_{i+}$ is the positive key representation, $k_{i-}^j$ are the negative key representations, $N$ is the number of queries, $K$ is the number of negative samples, and $\tau$ is the temperature parameter. This loss function forces the positive pairs to come closer and negative pairs are pushed further apart, thereby learning a representation that distinguishes different objects.

MoCo has shown competitive results in various tasks, including image classification, object detection, and semantic segmentation.

In our experiments MoCo outperformed BOYL, and was the visual encoder that we finally used in DiffClone.

\subsection{Agents for Policy Learning}
This part of the algorithm learns the policy, which maps the observation to the corresponding action. Here we consider behavioral cloning, and other offline RL algorithms to efficiently learn the action given a representation of the state of the robot arm.
\subsubsection{Imitation Learning via Behaviour Cloning}
Imitation Learning via Behaviour Cloning (BC) [\cite{BC}] is a machine learning approach where an agent learns to perform tasks by mimicking expert demonstrations. The objective in BC is to learn a policy \( \pi \) that maps states \( s \) to actions \( a \). This is often achieved through supervised learning by minimizing the discrepancy between the agent's actions and the expert's actions. Mathematically, this can be formulated as:

\begin{equation}
\min_{\pi} \sum_{(s,a) \in D} L(\pi(s), a)
\end{equation}

where \( D \) represents the dataset of expert state-action pairs, \( \pi(s) \) denotes the action predicted by the policy for state \( s \), \( a \) is the corresponding expert action, and \( L \) is a loss function, such as mean squared error, that measures the difference between the predicted action \( \pi(s) \) and the expert action \( a \). The policy is trained to replicate the expert behaviour as closely as possible. This in turn poses a major bottleneck, behaviour cloning is very brittle to generalization to new situations and can't handle the variability in the demonstrations.

We experimented with Behaviour Cloning, fine-tuned on each visual representation baseline to act as the agent. A quasi-open loop approach is used to predict action sequences of n-steps where n is a hyperparameter. In the TOTO-baseline, the value of n is kept at 50, and we used the same setup in our experiments.

\subsubsection{Visual Imitation via Nearest Neighbors (VINN)}

Visual Imitation through Nearest Neighbors (VINN) [\cite{VINN}] is a framework for visual imitation learning that decouples representation learning from behaviour learning. The VINN framework consists of two decoupled parts: training an encoding network on offline visual data using BYOL and querying against the provided demonstrations for a non-parametric locally weighted Nearest-Neighbor Regression based action prediction.

The nearest neighbors of the encoded input are found from the set of demonstration embeddings. The algorithm implicitly assumes that a similar observation must result in a similar action. Thus, once the k nearest neighbors of the query are found, the next action is set as a weighted average of the actions associated with those k nearest neighbors. This is done by performing nearest neighbors search based on the distance between embeddings, and then setting the action as the Euclidean kernel weighted average of those examples’ associated actions:

\begin{equation}
\hat{a} = \frac{\sum_{i=1}^{k} \exp(-\|e - e^{(i)}\|^{2}) \cdot a^{(i)}}{\sum_{i=1}^{k} \exp(-\|e - e^{(i)}\|^{2})}
\end{equation}

where $\hat{a}$ is the predicted action, $e$ is the encoded input, $e^{(i)}$ is the $i$th nearest neighbor, and $a^{(i)}$ is the action associated with the $i$th nearest neighbor.

\subsubsection{Offline RL Methods}
We also experimented with two standard offline reinforcement learning methods, namely Implicit Q-Learning (IQL) [\cite{IQL}] and Decision Transformer(DT) [\cite{DT}].

\textbf{IQL (Implicit Q-Learning)}: IQL is an offline reinforcement learning approach that estimates Q-values for state-action pairs without explicit policy optimization. It minimizes the Bellman residual while implicitly regularizing to avoid overestimation of unseen pairs. The objective is given by:
\begin{equation}
    \min_Q \mathbb{E}_{(s, a, r, s') \sim \mathcal{D}} \left[ \left( Q(s, a) - \left( r + \gamma \max_{a'} Q(s', a') \right) \right)^2 \right]
\end{equation}
where \( \mathcal{D} \) represents the dataset of transitions.

\textbf{Decision Transformer}: This method applies transformers to reinforcement learning, framing return-conditioned policy optimization as sequence modeling. The model predicts actions given a target return and past states and actions. The formulation is:
\begin{equation}
    \text{maximize } \mathbb{E}_{(G, s, a) \sim \mathcal{D}} [\log P(a | G, s)]
\end{equation}
where \( P \) models the probability and \( \mathcal{D} \) is the dataset.

\section{DiffClone: The Proposed Framework}

Behaviour cloning methods being a reward-free setup and being more data-efficient than RL methods, directly learn to mimic expert demonstrations, which might be better suited for these complex planning-based manipulation tasks without the need for exploration. Offline RL methods are also harder to train and optimize, particularly in sparse reward setups like robot manipulation. As a result in complex test scenarios having different objects, positions, and lighting conditions than the training scenarios, reward-free learning methods like BC tend to perform better particularly if the dataset is diverse and well-made with abundant optimal trajectories covering a variety of scenarios [\cite{kumar2022prefer}], all of which are satisfied by the TOTO dataset. 

We tried several experiments with the baseline visual encoders and the agent policies mentioned in the previous section. The experiments indeed justified our hypothesis, we found that Behaviour Cloning agents give superior results in comparison to other offline RL Methods such as IQL and Decision Transformer in complex manipulation tasks. This claim was further validated by the benchmark results of the TOTO paper and thus we focused on methods and architectures that would further improve the results of Behaviour Cloning. 

Typical methods in improving behaviour cloning are done by either looking more into the future (increasing horizon length) or by using better sequence modeling architectures such as RNNs, LSTMs etc. Diffusion models have proven successful in capturing complex distributions and efficiently preserving the multi-modality of the distributions they model. These claims were validated by the results achieved on various standard tasks using diffusion policy-based BC as mentioned in the work by \cite{diffpolicy}. We found that diffusion-policy performs better than other architectures for BC in our simulator environment as well.

\subsection{Data Preprocessing}
We use a MoCo finetuned ResNet50 [\cite{ResNet}], as a visual backbone which is provided by the authors of the TOTO paper.

For behaviour cloning, we found that using only the best trajectories(high-reward) trajectories led to more rewards and a better policy than using the whole dataset. Therefore, we restrict our data to only high-reward trajectories. Further after passing the image through our visual encoder, we append the current joint-states of the arm to get the observation.
State normalization is used for its capacity to enhance the stability of the taught policy and increase performance in several offline reinforcement learning benchmark tasks [\cite{fujimoto2021minimalist}]. The mean and variance of the states are calculated from dataset D. These parameters are then utilized for both training and testing. 
Therefore, we normalize the observation using the parameters, as mentioned above, before passing it to the agent.

\subsection{Diffusion Policy for Robot Behaviour}
The Diffusion Policy [\cite{diffpolicy}] introduced a novel approach for generating robot behaviour using a conditional denoising diffusion process. This method significantly outperforms existing robot learning methods across various benchmarks.

\begin{figure}[htp]
    \centering
    \includegraphics[width=10cm]{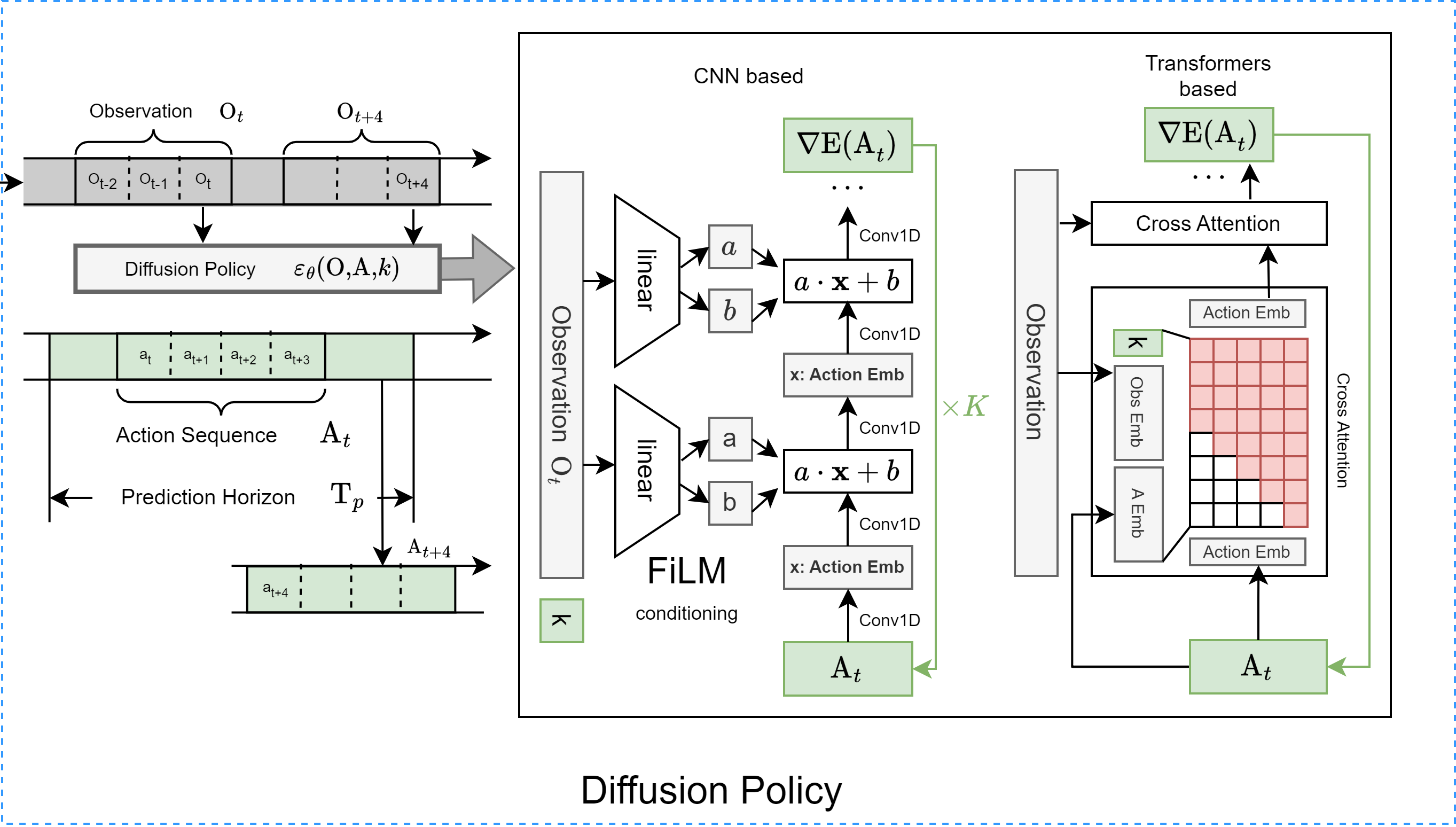}
    \caption{Diffusion Policy: A generative model that takes input the latest \( T_o \) observations \( O_t \) and predicts \( T_a \) subsequent actions \( A_t \), at each time step \( t \). In the CNN variant, it uses Feature-wise Linear Modulation (FiLM) for conditioning at each convolution layer \cite{film}. The Transformer-based approach \cite{attention} passes observation embeddings through a causally masked decoder with multi-head cross-attention.}
    \label{fig:g}
\end{figure}

\begin{itemize}

\item \textbf{Denoising Diffusion Probabilistic Models (DDPMs):} These models form the basis of the diffusion policy, expressed as:
    \begin{equation}
        x_{t-1} = \alpha_t x_t - \gamma_t \epsilon_\theta(x_t, t) + \mathcal{N}(0, \sigma_t^2 \mathbf{I}),
    \end{equation}
    where $x_t$ represents the state at timestep $t$, $\alpha_t$, $\gamma_t$, and $\sigma_t$ are time-dependent coefficients, and $\epsilon_\theta$ is the noise prediction network [\cite{DDPM}].

\item \textbf{Training Process:} The training involves minimizing the mean squared error between the actual and predicted noise, formulated as:
    \begin{equation}
        L = \text{MSE}(\epsilon_k, \epsilon_\theta(x_0 + \epsilon_k)).
    \end{equation}

\item \textbf{Adaptation for Visuomotor Policy Learning:} The formulation for visuomotor policy learning modifies the DDPM to represent robot actions and conditions the denoising process on input observations. This is expressed as:
    \begin{equation}
        A_{t_k-1} = \alpha(A_{t_k} - \gamma \epsilon_\theta(O_t, A_{t_k}, k) + \mathcal{N}(0, \sigma^2 \mathbf{I})),
    \end{equation}
    where $A_{t_k}$ denotes the action at time step $k$, and $O_t$ represents the input at time $t$.

\item \textbf{Noise Schedule:} The noise schedule, critical for capturing the characteristics of action signals, is defined by the Square Cosine Schedule [\cite{nichol2021improved}].

\end{itemize}

The Diffusion Policy method, leveraging Denoising Diffusion Probabilistic Models (DDPMs), excels in action gradient optimization through an iterative refinement process, crucial for robotic control tasks. This iterative refinement is mathematically represented in the equation above. 
The process utilizes the gradient of the log probability of the action given an observation,                     
\[ A_{t-1} = A_t - \eta \nabla_{A_t} \log p_\theta(A_t | O_t), \]
where \( \eta \) is the step size. This method is particularly effective in handling multimodal action distributions, a common challenge in robotic tasks. The approach incorporates elements of Stochastic Langevin Dynamics, 
\[ A_{new} = A_{old} + \eta \nabla_{A_{old}} \log p(A_{old}|O) + \mathcal{N}(0, 2\eta \mathbf{I}), \]
blending exploration and exploitation to refine action choices, demonstrating significant improvements in action prediction and execution in complex, multi-modal environments.

We use the policy's capacity to forecast action sequences in high-dimensional spaces and receding-horizon control to accomplish robust execution. This architecture enables the policy to consistently adjust its actions in a closed-loop fashion while ensuring temporal action consistency, so striking a balance between long-horizon planning and responsiveness.

Overall, the Diffusion Policy utilizes the effectiveness of DDPMs in visuomotor policy learning, and the action gradient-guided exploration of state space to achieve the best-performing agent, demonstrating remarkable improvements over existing offline and imitation methods, in handling multimodal action distributions and ensuring robustness and stability in training.

This Diffusion Policy is used on the offline TOTO dataset with a MoCo-finetuned ResNet50 as a visual backbone along with the data pre-processing and augmentations mentioned above as our final agent policy. We call it DiffClone.

\begin{algorithm}[htbp]

{\caption{DiffClone: Our Proposed Framework}}
{% contents
\begin{enumerate}
\item Prepare previously-collected data-set (here i is trajectory number and j is time-step in trajectory) D =$\{ x_{ij}$(image),$o_{ij}$(joint-state),$ a_{ij}$(action)$\}$
\item Initialize visual encoder $\phi,$ agent $ \theta$, and set the horizon number to be H
\item Pass each $x_i$ through the visual encoder $\phi$, to get image embeddings $e_{ij}$
\item Concatenate $o_{ij}$ to corresponding image embedding $e_{ij}$ to get observation $s_{ij}$
\item Perform state normalization over D for observations and actions
\item For training:  For $m=1$ to maximum number of epochs
    \begin{enumerate}
      \item Sample mini-batch of K transitions from D = $\{ x_{i},o_{i}, s_{i}, a_{i}\}$
      \item For $i=1$ to $K$
        \begin{enumerate}
        \item We have $s_{i}$ as the current observation, and $a_{i} = \{a_{i,j} | j = 0...H-1\}$ where $a$ denotes the next H actions taken from observation $s_{i}$
        % \item sample random noise_level_t (1,T)
        \item Sample random noise\_level\_t from (1, \, T)
        \item $a' = $ add\_noise($a_{i},\, $noise\_level\_t)
        \item $\gamma = \epsilon_{\theta}(a'\,,$noise\_level\_t,$\,s_{i})$ \; Predict added noise
        \item $L_{\theta} \leftarrow \,$compute\_loss($\gamma,\,a'-a_{i}\,$)
        \item Update $\theta$
      \end{enumerate}
    \end{enumerate}
\item For inference: For sample s=$\{ x,o\}$ in $D_{test}$
    \begin{enumerate}
        \item Pass $x$ through the visual encoder $\phi$ to get embedding $e$
        \item Concatenate $o$ to $e$ to obtain $s$, where s is current observation 
        \item $a = $ initialise\_with\_noise()
        \item For t in reversed(time\_steps(0,T)):
        \begin{enumerate}
            \item $\gamma = \epsilon_{\theta}(a\,,$noise\_level\_t,$\,s)$
            \item $a =$ remove\_noise($a,\, \gamma,\, $noise\_level\_t)
        \end{enumerate}
    \end{enumerate}
\end{enumerate}
}
\label{alg:DiffClone}
\end{algorithm}
\begin{figure}[htp]
    \centering
    \includegraphics[width=15cm]{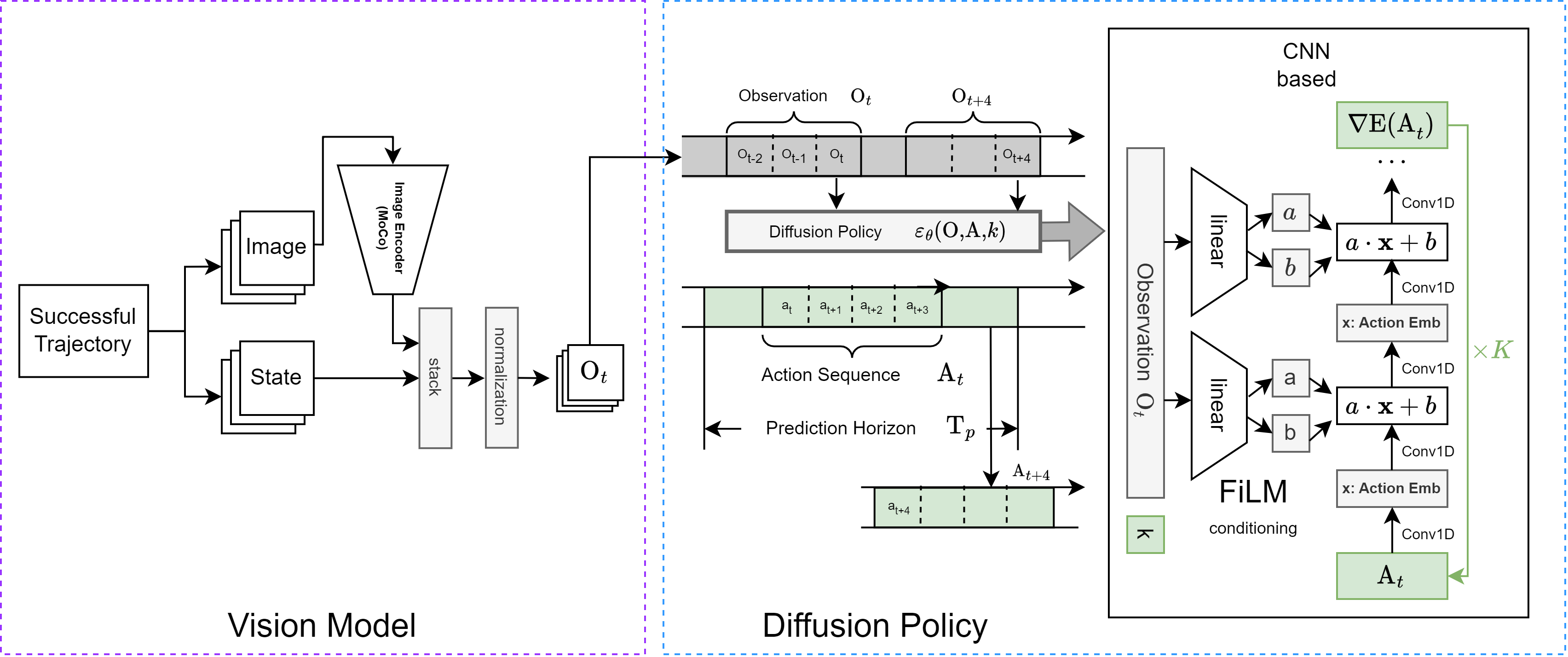}
    \caption{Schematic Model of our proposed DiffClone Framework}
    \label{fig:galaxy}
\end{figure}

\section{Experiments and Results}

We tried multiple experiments and ablations with the visual representation and agent policy, before arriving at our final agent- DiffClone. The following sections give a brief overview of our initial experimentation, followed by our results using DiffClone.

\begin{figure}[H]
    \centering
    \includegraphics[width=15cm]{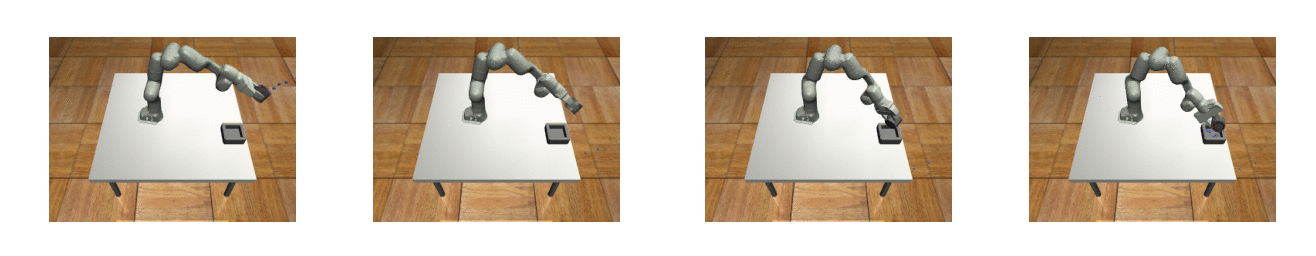}
    \caption{Agent learning policy gradually using DiffClone}
    \label{fig:galaxy_1}
\end{figure}

\subsection{Experiments with Visual Representation}
To further highlight the information of the state of the robot and the dynamics of successive steps, we tried to map the delta of successive embeddings to the delta of their successive states using an MLP, and this served as a contrastive objective. This gave a small increase in rewards and retained the success rate with behaviour cloning. We also tried adding a small objective of goal conditioning, where we approximated the goal to the final state of the arm and then tried mapping each embedding to its corresponding goal state. We also attempted to use the depth images given to extract a mask using off-the-shelf models.
These experiments did not lead to significant improvement in rewards or success rates.

\subsection{Experiments with Agent Policy}
We tried various offline RL algorithms such as Q-learning, implicit Q-Learning, and trajectory prediction transformers, but these did not yield better results than the ones we got from direct behaviour cloning.
We then focused on experimenting with various sequence modeling architectures such as RNNs, LSTMs, and Diffusion Policy. Diffusion policy gave significantly better results than plain behaviour cloning in simulation. In the next section, we will discuss the ablations and hyper-parameter tuning we did with diffusion policy to arrive at our best submission.

We performed ablation studies by varying architecture and hyper-parameters such as de-noising steps, sub-sampling period, horizon length, open loop-closed loop inference, and the architecture used for the DDPM. These experiments are discussed in the following section.

\subsection{Experiments with Diffusion Policy}

\subsubsection{Architectural Choice}
% Cite motion diffusion models paper
Inspired by Motion Diffusion Models [\cite{tevet2022human}], a transformer encoder architecture was used for the noise prediction network. We first concatenated the current observation with the sequence of actions and used an Attention mechanism to learn the conditional mappings and the temporal change across the sequence of actions. This method did not successfully learn the diversity of the actions, resulting in a mode collapse. As a possible solution, we could have experimented with a Diversity Loss across batches, but we were unable to experiment further with the due to compute constraints. 
Also, Transformer encoders are known to be very sensitive to hyperparameters. Hence, we shifted to 1D-Conditional U-Net architectures for the same task, which was able to capture the distribution. We experimented with directly predicting the denoised sample instead of the added noise, but it didn't give any improvement. We also could not experiment with training the encoder + diffusion policy in an end-to-end fashion due to compute limitations. This approach may give better results.
\begin{table}[H]
    \centering
    \begin{tabular}{ccc}
        \textbf{Model} & \textbf{Mean Reward} & \textbf{Success Rate}\textbf{(}\textbf{\%}\textbf{)}\\ \hline \hline
         \textbf{DiffClone (U-Net based)} (\textbf{Ours})&  \textbf{51}& \textbf{92}\\
         DiffClone (Transformer based) (\textbf{Ours})&  5.6& 24\\
         MoCo + BC (\textbf{TOTO})&  20.33& 68\\ \hline
    \end{tabular}
    \caption{Our simulation evaluation on Pouring}
    \label{tab:my_label_2}
\end{table}

\begin{table}[H]
    \centering
    \begin{tabular}{ccc}
        \textbf{Model} & \textbf{Mean Reward} & \textbf{Success Rate}\textbf{(}\textbf{\%}\textbf{)}\\ \hline \hline
         DiffClone (130 Epochs) (\textbf{Ours})&  7.833& 33.33\\
         DiffClone (2000 Epochs) (\textbf{Ours})&  6.417& 33.33\\
         MoCo + BC (\textbf{TOTO})&  22.86& 72.2\\
         DiffClone (Subsampling 3) (\textbf{Ours})&  0& 0\\ \hline
    \end{tabular}
    \caption{Real Robot Testing on Pouring}
    \label{tab:my_label_1}
\end{table}

\begin{table}[H]
    \centering
    \begin{tabular}{ccc}
        \textbf{Model} & \textbf{Mean Reward} & \textbf{Success Rate}\textbf{(}\textbf{\%}\textbf{)}\\ \hline \hline
         DiffClone (130 Epochs) (\textbf{Ours})&  6.91& 58.30\\
         DiffClone (600 Epochs) (\textbf{Ours})&  2& 25\\
         MoCo + BC (\textbf{TOTO})&  7.42& 83.3\\ \hline
    \end{tabular}
    \caption{Real Robot Testing on Scooping}
    \label{tab:my_label}
\end{table}

\subsubsection{Hyperparameters}
We used the DDPM scheduler from Diffusers Library [\cite{diffusers}] during training with 50 time steps. When the same scheduling is used during inference, it may lead to inconsistent and broken motion of the robot arm due to high latency. We can use DDIM with fewer time steps to resolve this issue and improve latency.

Sub-sampling of actions from training data proved to be an influential factor in training stability. Higher sub-sampling rates (> 3) resulted in the loss of information, which led to sub-optimal policies and mode collapse in some instances. On the other hand, lower sub-sampling rates increase the number of data points used in training, thus allowing us to cover the entire dataset, but as a trade-off resulted in increased training. We found that in simulation, a sub-sampling period of 1 gives the best results.

We use a prediction horizon length of sixteen but recalculate after every eight action steps. This is a trade-off between goal-aware long-horizon trajectories and the responsiveness of the policy, as mentioned above in the methodology. We have added the hyperparameter table in the appendix.

Our algorithm performed extremely well and significantly better than TOTO-baselines in our evaluation in simulation but failed to do so in real-world testing. This may be due to the fact that they do not generalize well or are very sensitive to hyper-parameters that need to be readjusted for efficient and successful real-world implementation. We plan to explore and re-evaluate this concern as a part of our future work.

\section{Conclusion}
In this paper, we introduce DiffClone, a diffusion-based behavior cloning agent that performs complex robot manipulation tasks from offline data. Our method captures complex distributions and efficiently preserves its multi-modality, thus solving the offline RL problem in an efficient and robust fashion. To this end, we adopted a state-of-the-art diffusion-policy-based approach along with a MoCo fine-tuned Resnet50 visual backbone. In our evaluation, we found that our policies achieved high scores compared to the established baselines in the simulation. At the same time, we observed that diffusion policies are very susceptible to changes in hyper-parameters, such as the number of de-noising time steps and sub-sampling period. In our future work, we plan to implement DDIM for improved latency during inference and explore regularisation methods such as KL regularisation to enable a more robust transfer to real-time environments. We plan to further explore methods for the efficient transfer of our algorithm from simulation to the real world. The code of our work is open-sourced at the following link: \url{https://github.com/sirabas369/DiffClone.git}, the project website is as follows: \url{https://sites.google.com/view/iitkgp-nips23toto/home}, it also contains a few working videos of our trained policies.

\subsection{Acknowledgement}
We would like to take this opportunity to thank the organisers of TOTO benchmark challenge for giving us an opportunity to explore this paradigm and for open-sourcing their code and dataset. We would also like to thank the authors of Diffusion Policy, for open-sourcing their implementation of the CNN-based diffusion policy, and  Dr. P. P. Chakrabarti (Professor, Dept. of CSE, IIT Kharagpur), for his invaluable inputs and guidance.

\bibliographystyle{unsrtnat}
\bibliography{main}

%%%%%%%%%%%%%%%%%%%%%%%%%%%%%%%%%%%%%%%%%%%%%%%%%%%%%%%%%%%%

%Appendix is kept commented below. Unless necessary
\appendix
% Appendix 1
\section{Appendix}

\subsection{Hyperparameters}
\begin{table}[H]
\centering
\small
\renewcommand{\arraystretch}{1.3}
\begin{tabular}{ll}
\hline
\textbf{Hyperparameter}  & \textbf{Value}\\ \hline \hhline{}
Batch Size & 128 \\ 
%Epochs & 4 \\ \hline
Prediction Horizon & 16\\ 
Execution Horizon & 8\\ 
Sub-sample Period & 1 \\ 
Denoising time-steps & 50 \\ 
Action Dimension & 7 \\
Learning rate & 1e-4 \\ \hline

\end{tabular}
\caption{Hyperparameters used for the MoCo DiffClone model}
\label{hyperpTable}
\end{table}

\subsection{Other Experiments}
    \begin{table}[H]
    \centering
    \begin{tabular}{ccc}
        \textbf{Model} & \textbf{Mean Reward} & \textbf{Success Rate}\textbf{(}\textbf{\%}\textbf{)}\\ \hline \hline
          MoCo + BC (\textbf{TOTO})&  20.33& 68\\
         $\delta$ observations -$\delta$ actions(transformer encoder) &  15& 36\\
         $\delta$ observations -$\delta$ actions(MLP )&  19& 72\\
         $\delta$ observations -$\delta$ joint$\_$states(MLP)&  31.33& 68\\
         $\delta$ observations -$\delta$ joint$\_$states(transformer encoder)&  11& 32\\ \hline
    \end{tabular}
    \caption{Our simulation experiments on Pouring }
    \label{tab:my_label}
    \end{table}

    \subsection{Training and Test Plots}

\begin{figure}[htp]
    \centering
    \includegraphics[width=15cm]{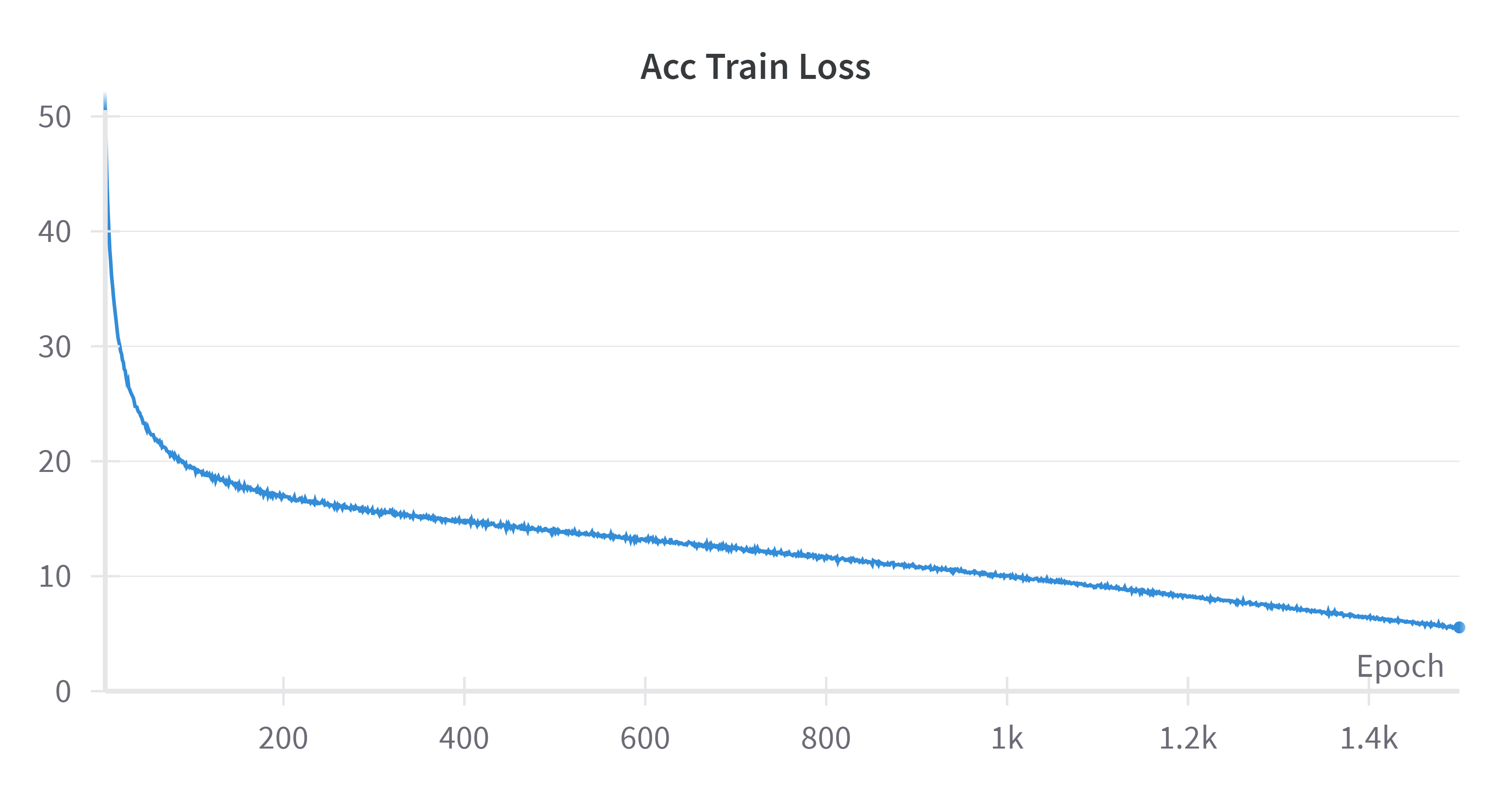}
    \caption{Pouring Training Loss Plot}
    \label{fig:galaxy}
\end{figure}

\begin{figure}[htp]
    \centering
    \includegraphics[width=15cm]{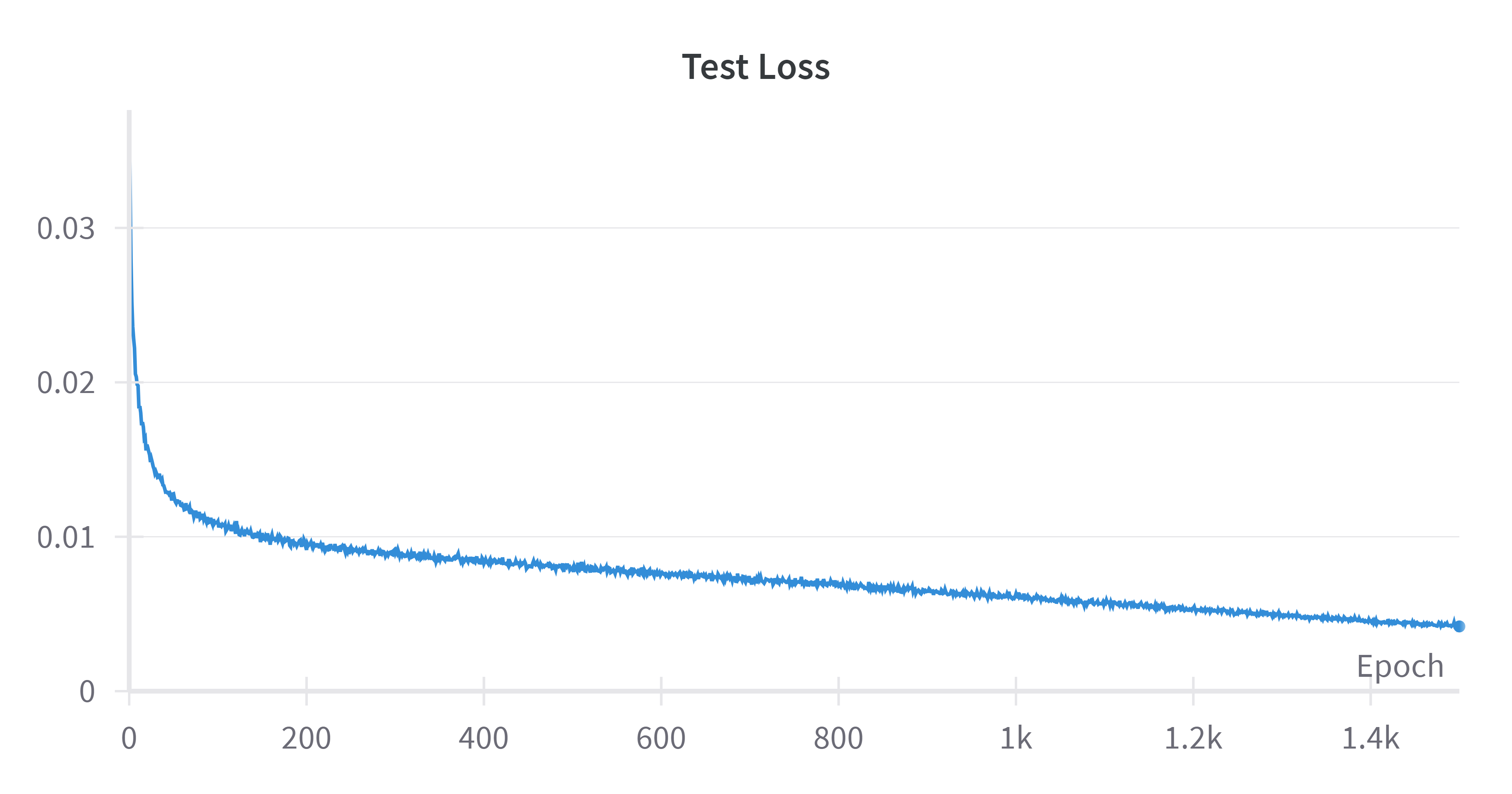}
    \caption{Pouring Simulation Test Loss Plot}
    \label{fig:galaxy}
\end{figure}

\begin{figure}[htp]
    \centering
    \includegraphics[width=15cm]{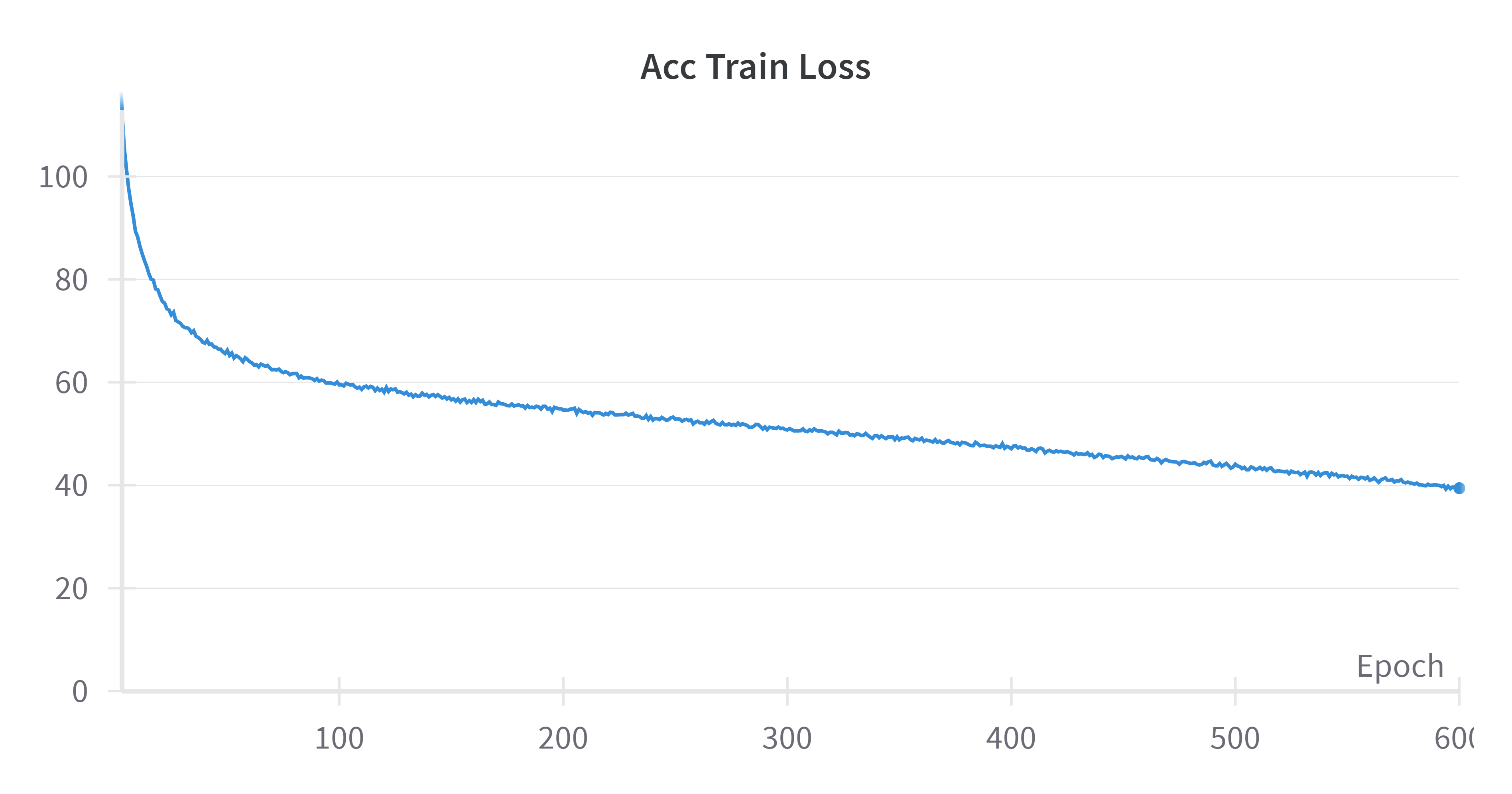}
    \caption{Scooping Training Loss Plot}
    \label{fig:galaxy}
\end{figure}

\end{document}